\documentclass[runningheads]{llncs}
\usepackage{graphicx}
\usepackage{amsmath,amssymb} % define this before the line numbering.
\usepackage{makecell}
\usepackage{caption} 
\usepackage{color}
\usepackage{authblk}

\begin{document}
%===========================================================

\title{Deep Reader: Information extraction from Document images via relation extraction and Natural Language} % Replace your paper's title here
\titlerunning{Deep Reader: Information extraction from Document images} % Replace an abstracted version of your paper's title here

%===========================================================

%\author{First Author\inst{1}\orcidID{0000-1111-2222-3333} \and
%Second Author\inst{2,3}\orcidID{1111-2222-3333-4444} \and
%Third Author\inst{3}\orcidID{2222--3333-4444-5555}}

%\authorrunning{Vishwanath, Rohit, Gunjan, Arindam, Monika, Lovekesh, Gautam, Ashwin}
\authorrunning{Vishwanath D et al}
%===========================================================
\author{Vishwanath D\inst{1 *}, Rohit Rahul\inst{1 *}, Gunjan Sehgal\inst{1}, Swati\inst{1}, Arindam Chowdhury\inst{1}, Monika Sharma\inst{1}, Lovekesh Vig\inst{1}, Gautam Shroff\inst{1}, Ashwin Srinivasan\inst{2}\\}

\institute{ TCS Research, New Delhi\inst{1}, BITS Pilani, Goa Campus\inst{2} \\
\texttt{ \{vishwanath.d2, rohit.rahul, sehgal.gunjan, j.swati, chowdhury.arindam1, monika.sharma1, lovekesh.vig, gautam.shroff \}@tcs.com}\\
\texttt{ashwin@goa.bits-pilani.ac.in}
}

\maketitle
%===========================================================
\begin{abstract}
Recent advancements in the area of Computer Vision with state-of-art Neural Networks has given a boost to Optical Character Recognition (OCR) accuracies. However, extracting characters/text alone is often insufficient for relevant information extraction as documents also have a visual structure that is not captured by OCR. Extracting information from tables, charts, footnotes, boxes, headings and retrieving the corresponding structured representation for the document remains a challenge and finds application in a large number of real-world use cases. In this paper, we propose a novel enterprise based end-to-end framework called DeepReader which facilitates information extraction from document images via identification of visual entities and populating a meta relational model across different entities in the document image. The model schema allows for an easy to understand abstraction of the entities detected by the deep vision models and the relationships between them. DeepReader has a suite of state-of-the-art vision algorithms which are applied to recognize handwritten and printed text, eliminate noisy effects, identify the type of documents and detect visual entities like tables, lines and boxes. Deep Reader maps the extracted entities into a rich relational schema so as to capture all the relevant relationships between entities (words, textboxes, lines etc) detected in the document. Relevant information and fields can then be extracted from the document by writing SQL queries on top of the relationship tables. A natural language based interface is added on top of the relationship schema so that a non-technical user, specifying the queries in natural language, can fetch the information with minimal effort. In this paper, we also demonstrate many different capabilities of Deep Reader and report results on a real-world use case.

\let\thefootnote\relax\footnote{* Equal contribution authors}

\end{abstract}
%===========================================================
\section{Introduction}
With the proliferation of cameras and scanning software on mobile devices, users are frequently uploading a variety of scanned images of documents such as invoices, passports and contracts to application servers on the cloud. Currently, much of the processing of these documents is at least partially done manually owing to the critical nature of the transactions involved. However, with recent advancements in deep learning for vision applications, it has become possible to further automate this process. The problem falls under the realm of information extraction from images and has been a longstanding research problem for vision researchers \cite{peanho2012}. 

While OCR accuracies have significantly improved, thanks to advancement in deep learning, these alone are insufficient for effective extraction of visual information from scanned documents. Most documents have a rich set of visual entities in the form of tables, text-boxes, blocks, charts, and arrows. Until recently, vision algorithms were not powerful enough to accurately identify and extract these visual entities, which resulted in errors being propagated downstream in the extraction pipeline. Real world documents often have a combination of both handwritten and printed text which makes text harder to localize and identify. Additionally, the scanning process can often introduce noise in the documents which confuses the text recognition algorithms. Therefore, any real world deployment has to address these vision challenges in order to be effective. 

In addition to address the vision problems mentioned above, there are challenges involved in understanding the complex visual structure between the entities in the document. The visual relationships between the different entities detected in an image are often critical to understanding and reasoning over the information present prior to extraction. For example a text label might only make sense if viewed in the context of the entity it is connected to via an arrow. 

Humans utilize a lot of universal background knowledge while reading documents, and this needs to be incorporated into any extraction engine. For example, we know that an address comprises of a city and country name. This knowledge needs to be captured and embedded into the system. Also, very often incorporation of domain knowledge or business rules can often boost the extraction performance and enable validation and correction of extracted data. 

In this paper we present Deep Reader, a platform for information extraction from images which attempts to incorporate these salient relationships prior to information extraction. The platform utilizes state of the art deep learning based vision algorithms for denoising the image documents and identifying entities such as tables, printed and handwritten text, text blocks, boxes and lines. The spatial relationships between these entities are then recorded and used to populate a high level relational schema. The relational schema in question is designed to be as generic and exhaustive as possible. Users are allowed to filter and retrieve relevant data from the populated schema either via SQL queries, or via a conversational interface. This allows users to incorporate domain specific information and business rules in their queries, for example SWIFT address must comprise of 8 alphanumeric characters, drawer name always occurs before drawee etc. Once the user is satisfied with the results of their queries, the queries are stored and may be applied to other documents sharing the same template. 
 
\section{Related Work}
Extracting text from images have been an active field of research for several decades. With the advent of deep neural networks, OCR engines have become very powerful with opensource offerings like Tesseract \cite{smith2007overview}, and cloud API based solutions like Google Text Vision API. Interpreting documents with a relatively simple textual layout and good quality scans is now reasonably straightforward thanks to these advancements. However, when dealing with documents following several different templates with diverse visual layouts, retrieving semantically accurate information can be very challenging. There has been extensive line of work towards solving this problem.

 Yshitani [18] developed an information extraction system wherein a document instance is matched with a set of pre-stored models which define categories of documents. The extracted document text is compared against a pre-compiled list of keywords and their spelling variants. The results are matched with each document in a database of word models and their logical relationships. 

Cesarini \cite{cesarini98} requires user to build a conceptual model of the document, which is then used to match, interpret and extract contents from the document. The work places more emphasis on the classification and organization of the documents rather than extraction of key fields. 
Both Cesarani \cite{cesarini98} and Peanho \cite{peanho2012} build an Attribute Relational Graph based representation of the document to capture relationships between entities in an image, however their system relies on considerable expertise from user to create suitable template document models.

\cite{rusinol2013field} propose a system for extracting field information from administrative documents using a document model for extracting structural relationships between words. As the system processes more documents, it refines its structural model. More recently, \cite{hammami2015one} proposed a technique to extract data from colored documents by extracting rectangles and modeling the topological relationship between them as a graph. The technique was generalized to extraction from new cases based on topological and context similarity. An Entity Recognition model was proposed in  \cite{kooli2015semantic}, using geometrical relationships between entities.  \cite{aldavert2017automatic} propose a system for automatic separation of static and variable content in Administrative Document Images using a probabilistic model for determining whether a pixel is static or dynamic.

A relevant problem for extraction is that of document template identification for which several techniques have been proposed. Bruel developed a matching technique that was more robust to noise in the template instance \cite{breuel2001}, and subsequently to capture the layout of a complex document \cite{breuel2003}. In [9], the authors generate a hierarchical representation of documents and templates using XY-trees and utilize tree edit distance to identify the correct template for a new instance. Hamza\cite{hamza2007case} utilize a bottom-up aproach which involves keyword extraction followed by clustering to generate high level structures. The document is represented as a graph with these high level structures as nodes and edges indicative of spatial relationships. Documents are mapped via graph probing, which is an approximation of graph edit distance.   
Schultz et al~\cite{schulz2009seizing} have developed an invoice reading system, where documents are again represented by a graph but nodes are split into key-nodes and data-nodes, corresponding to keywords and values (numbers, dates etc). Graph probing is used to find the correct template. Extraction is done by examining the nearby words of every data-node.  

In fact, in many of these works approximate  solutions  to  known  hard  problems  (such  as  graph distance, tree edit distance and maximum cliques) have been used to match an input document with a document model. In DeepReader, we utilize a combination of visual structure similarity  and textual similarity (captured by a deep Siamese network) to classify documents to the appropriate template. 

Information Extraction from documents remains an open problem in general and in this paper we attempt to revisit this problem armed with a suite of state of the art deep learning vision APIs and deep learning based text processing solutions. These include utilization of generative adversarial networks\cite{goodfellowij} for image denoising, Siamese networks\cite{koch2015siamese} for document template identification and sequence to sequence models for handwritten text recognition \cite{chowdhury2018efficient}. DeepReader also utilizes low-level vision routines that are extremely useful for identifying text based entities in an image and populating a pre-defined relational schema. This relational schema can then be queried using standard SQL. 

 The advent of sequence to sequence models has allowed for significant advancement in several text modeling tasks and these models play a crucial role in several state of the art conversational systems\cite{vinyals2015neural}. DeepReader utilizes a sequence to sequence model to provide users with the ability to query the system via a natural language conversational interface. The interface builds on prior work that utilizes seq2seq models for learning an NL to SQL query mapping\cite{yu2018typesql}  The motivation is to allow non-technical users to efficiently utilize the system for efficient data extraction from images.

%Do not use any additional Latex macros.

%------------------------------------------------------------------------- 

\section{DeepReader Architecture}

\begin{figure*}[ht!]
\begin{center}
%\fbox{\rule{0pt}{2in}\rule{0.9\linewidth}{0pt}}
\includegraphics[width=\linewidth,height=6cm]{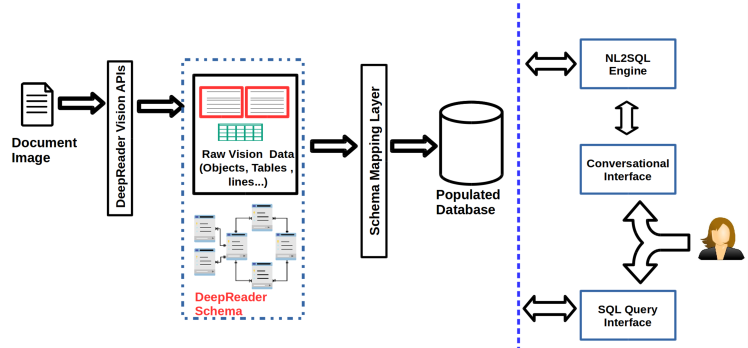}
\end{center}
\caption{Deep Reader Workflow for writing rules for a query}
\label{deepReader_arch}
\end{figure*}

DeepReader processes documents in several stages as shown in Figure \ref{deepReader_arch}. Initially a raw document image which may be blurred, noisy or faded is input to the system. A denoising suite is employed to clean the image prior to data extraction. The clean image is then sent for document identification to ascertain the correct template for the document. DeepReader vision APIs are subsequently employed for entity extraction followed by a schema mapping layer to populate a relational database that adheres to the DeepReader schema. The user may retrieve contents from the database either via SQL or via a Conversational Interface.

%---------------------------------------------------------------------------

\section{Computer Vision Components of Deep Reader}
%OCR for recognition of printed text: Agnostic to the underlying model, can utilize Google Vision API, Abby Fine Reader or Tesseract.
To pre-process a document for information extraction, a number of visual elements need to be identified. These include printed and handwritten text, text lines, clusters of text lines( also referred to as textblocks), boxes, and tables. DeepReader leverages a number of deep vision APIs in order to identify the different visual elements present in the document. Additionally, a document cleaning suite based on generative adversarial networks is developed and is a vital pre processing step as described below.

\subsection{Image De-Noising}
In this section, we address the issue of degradation in quality of images due to camera shake, improper focus, imaging noise, coffee stains, wrinkles, low resolution, poor lighting, or reflections. These kind of problems drastically affect the performance of many computer vision algorithms like text detection, OCR and localization. The objective here is to reconstruct high-quality images directly from noisy inputs and also to preserve the highly structured data in the images.

Text document images are markedly different from natural scene images as text documents contain more detailed information and are therefore more sensitive to noise. Thus, we propose a denoising method by utilizing generative adversarial networks (GANs)~\cite{goodfellowij}. Generative Adversarial Networks have been applied to different image-to-image translation problems, such as super resolution~\cite{ledig2017photo}, style transfer~\cite{li2016precomputed}, and product photo generation~\cite{bousmalis2017unsupervised}. In DeepReader, we have implemented conditional generative adversarial networks (cGANs) in which both the generator and discriminator are conditioned on extra information $\textbf{y}$ as described in \cite{zhang2017image}. In our case, the variable $\textbf{y}$ is represented by a class label i.e., the cleaned image. We have conditioned only on the discriminator by feeding $\textbf{y}$ as an extra input layer. The cGANs network is trained using the following minimax objective funtion:

\begin{equation}
\min_{G} \max_{D} E_{x\sim P_r} [log(D(x|y))] + E_{\widetilde{x}\sim P_g} [log(1 - D(\widetilde{x}|y))]
\end{equation}

where $P_r$ is data distribution and $P_g$ is model distribution defined by $\widetilde{x} = G(z)$, $z =P(z)$ and $z$ is one of the samples from the noisy images dataset.

\begin{figure*}[ht]
\centering 
\includegraphics[width=\linewidth]{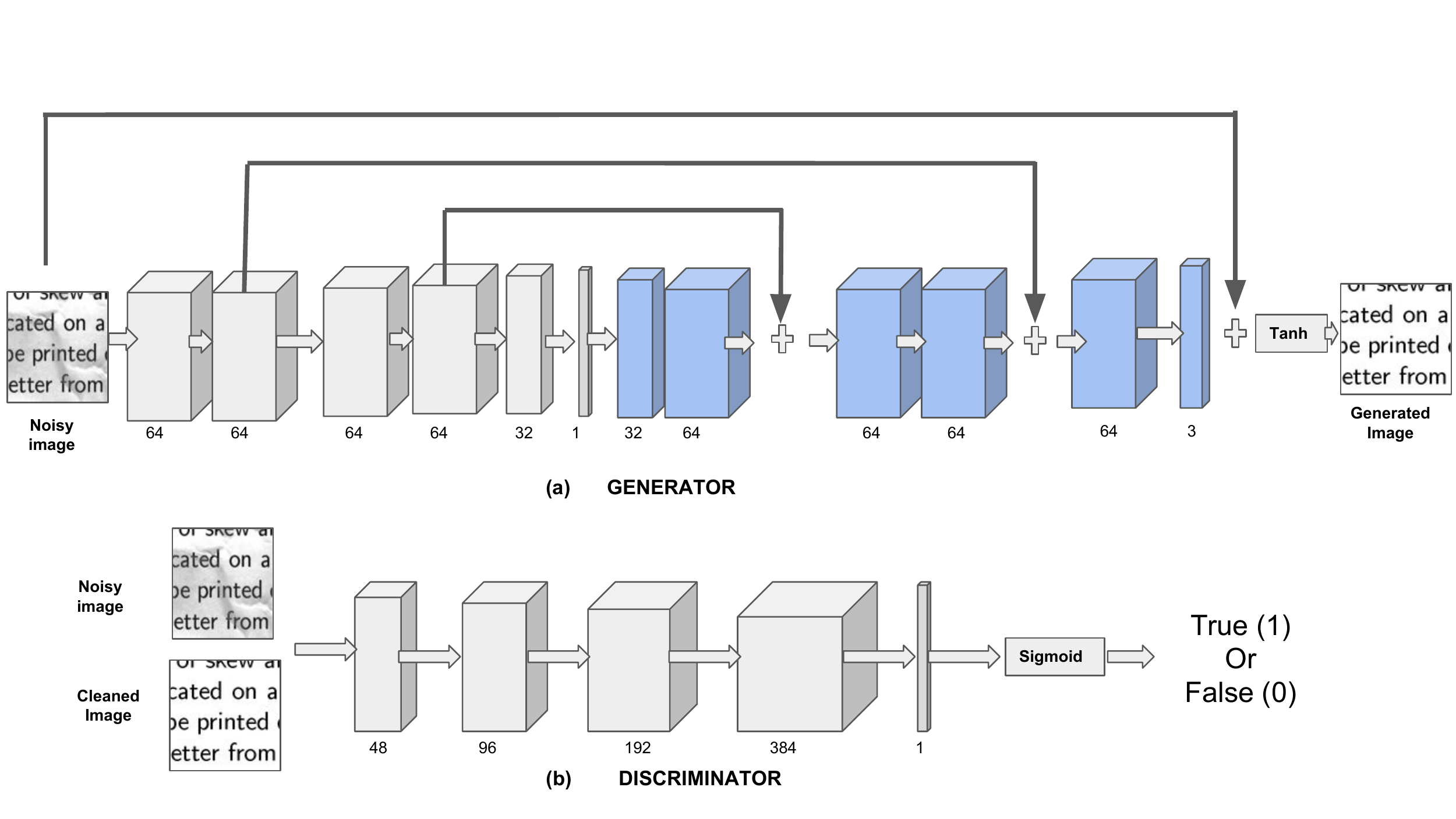}
\caption{The architecture of cGANs for denoising images. The upper figure shows the generator model and the lower figure shows the discriminator model. The generator model takes noisy images as input and generates a high-quality image while the discriminator takes both clean and noisy images and learns to differentiate between them. Reproduced from \cite{zhang2017image}.}
\label{fig:arch}
\end{figure*}

\textbf{Generator:} The architecture of our network is shown in Figure \ref{fig:arch}. The generator network consists of a combination of convolutional layers (represented by grey color blocks) and deconvolutional layers (represented by blue color blocks). Each of the blocks is followed by batch normalization and the activation used in the convolutional layer is PReLU while for deconvolutional layer, ReLU activation is employed. For all the layers, the stride is set to be 1. In order to maintain the dimension of each feature map to be the same as that of the input, we use zero padding. For efficient network training and better convergence performance, symmetric skip connections are used as shown in the generator network. 

\textbf{Discriminator:} The discriminator is used to classify each input image into two labels - real or fake. The discriminator network consists of convolutional layers each having batch normalization and PReLU activation. A sigmoid function is stacked at the top to map the output to a probability score between [0,1].

The loss function used for the cGAN network is same as define in \cite{zhang2017image}:

\begin{equation}
L = L_E + \lambda_a L_A + \lambda_p L_P
\end{equation}

where, $L_E$ is per-pixel loss function, $L_P$ is perceptual loss and $L_A$ is adversarial loss (loss from the discriminator). $\lambda_a$ and $\lambda_p$ are pre-defined weights for adversarial loss and perceptual loss, respectively.\\

\begin{figure*}[ht]
\centering 
\includegraphics[width=\columnwidth, height=20mm]{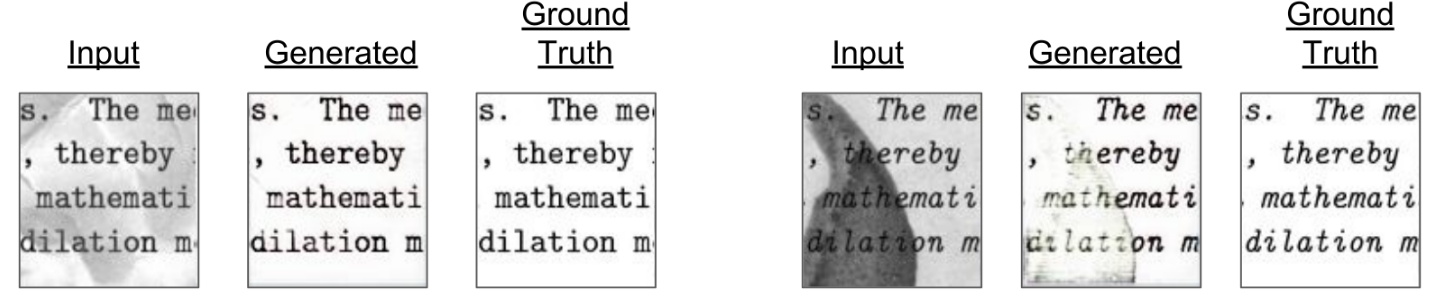}
\caption{Sample output after de-noising the image}
\label{denoisedoutput}
\end{figure*}

\textbf{Peformance evaluation:} We have evaluated our model on a public kaggle document dataset\footnote{https://www.kaggle.com/c/denoising-dirty-documents} hosted by~\cite{frank2010uci}. The performance measure used for the evaluation of the model are Peak Signal to Noise Ratio (PSNR) and Structural Similarity Index (SSIM)~\cite{wang2004image}. The value of PSNR and SSIM obtained are \textbf{17.85 dB} and \textbf{0.946}, respectively. Some of the output images obtained from generator model are shown in Figure \ref{denoisedoutput}.

%\begin{figure*}[ht]
%\centering 
%	\includegraphics{figure=denoisedoutput.pdf,width=\linewidth, height=70mm}
%\caption{Resultant images generated by generator model (shown in middle column of each row).}
%\label{denoisedoutput}
%\end{figure*}

\subsection{Document Identification}
One of the more pressing challenges in the pipeline is determining the correct template in which to classify the test document. We utilize a deep Siamese network for this~\cite{koch2015siamese}:

The base of our Siamese network consists of a traditional Convolutional Neural Network (CNN). It consists of two convolutional layers with 64 and 128 filter size respectively. Each convolutional layer operates on stride of 1 and is followed by a max pool layer of size $2 \times 2$, which is also operating at a stride of 1. All convolutional layers in our base CNN have Rectified Linear Units (ReLU) as an activation function. The output of the final max pool layer is flattened into a vector and is passed into the succeeding fully connected dense layer having 512 hidden units and sigmoid activation. This dense layer is followed by the computation of an energy function over the feature representations at the highest level. The Siamese network is learnt using a contrastive loss function~\cite{chopra2005learning}. We trained the siamese network on an image size of $500 \times 400$. While training, the optimizer used is Stochastic Gradient Descent (SGD) with learning rate of $10^{-2}$, Nestrov momentum value was set to $0.99$ and weight decay was set to $10^{-6}$. We evaluated the performance of our model on a dataset containing 9 types of bank documents and the model yielded an accuracy of $\textbf{88.89\%}$.

\subsection{Processing Handwritten Text}
 A major challenge in extracting useful information from scanned text documents is the recognition of handwritten text(HTR) in addition to printed text. It is an important problem for an enterprises attempting to digitize large volumes of handmarked scanned documents. Towards this end, we use the HTR system proposed in \cite{arindam_htr} which uses a convolutional feature extractor followed by a recurrent \textit{encoder-decoder} model for mapping the visual features to a set of characters in the image. A general overview of the model is provided in \cite{chowdhury2018efficient}.

\begin{figure*}[ht]
\centering 
\fbox{ \includegraphics[width=0.7\columnwidth, height=40mm]{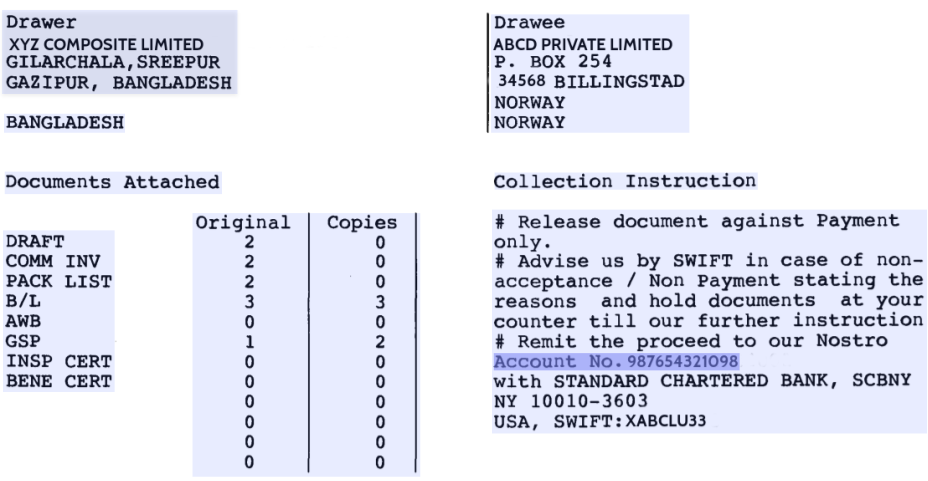} }
\caption{Sample Bank Document}
\label{fig:sample_city}
\end{figure*}

\section{Extracting different text entities from the document}
A document has many entities like words, lines, text blocks, and text boxes. The information of interest may involve one or several of these entities, which makes it imperitive to detect all of the entities present in the document.

\subsection{Detecting Textual Entities}
We utilize a combination of connected components, and spatial clustering to identify and cluster text characters in the image into words, lines and blocks. For scene text or colored images present in the document, we use the connectionist text proposal network (CTPN) \cite{tian2016detecting}. The objective is to get the location of the different textual entitites present in the image. 

\subsection{Entities}
The following high level entities were extracted:
\begin{enumerate}
\item \textbf{Page lines:} To get the page lines we perform horizontal clustering on the words based on the Euclidean Distance between connected component clusters. After we localize the page lines, each page line patch is sent through  a standard OCR engine (such as Tesseract, Google Vision or Abby FineReader) to get the text corresponding to that line. We noticed that sending smaller patches resulted in higher accuracy from the OCR engine. 

\item \textbf{Text Block:} A text block is a set of lines which begin at approximately the same $x$ coordinate and the vertical distance between them is not more than twice the height of the line. Fig \ref{fig:sample_city} shows a highlighted block. 

\item \textbf{Lines of the Text Block or Box:} The lines that lie inside a particular textblock or box are also identified seperately as block or box lines.

%\item \textbf{Lines for the Text Block:} The process of finding the lines for finding the text block is similar to finding the page lines in the image. The difference is that we consider the interspace distance between the words. for instance in fig 6 the words "Documents Attached" and "Collection Instruction" would be two seperate lines. To find the page lines we set the threshold T1 to the moving average to the height of the bounding box multiplied by 18. Afterwards, the text recognition is done by the google vision API which gives us the text corresponding to each lines.
\item \textbf{Boxes:} To find the Boxes, we first erode the image followed by the thresholding and inversion. After that we compare the area of each connected component with the area of its bounding box. If the area of the connected component is within a percent of the area of bounding box then we deem that connected component as a box. 
\end{enumerate}

\section{Deep Reader Schema}
Once all the entities are identified as mentioned in the above section, relations between the entities need to be populated and stored in the database. The corresponding database schema should be designed to facilitate the subsequent information extraction. All the entities are associated with their spatial coordinates and this information conveys the whereabouts of the neighbouring  text entities. This information is then  used to infer different logical and spatial relationships.

\begin{figure*}[ht]
\centering 
\includegraphics[scale=0.3]{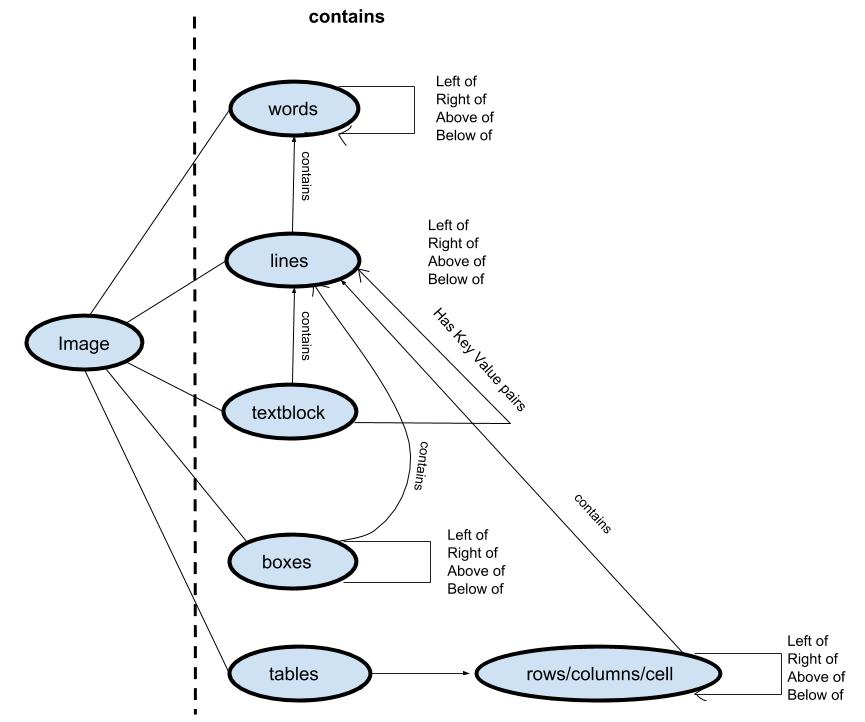}
\caption{Deep Reader Schema}
\label{schema}
\end{figure*}

The line entities identified from the vision components includes information about the line text, individual words in the line along with the line and word coordinates.  Using the coordinate position of words, DeepReader extracts words to the left/right/above/below of other words and maintains it in the schema. Similarly it uses the raw vision data to maintain the words and the lines in which they occur in the schema as a separate table. As shown in the figure \ref{fig:sample_city}, the word 'SREEPUR' will be detected by the vision components along with the associated coordinates. The word to the left is 'GILARCHALA', right is 'null', above is 'COMPOSITE' and below is 'BANGLADESH'. In this way deep reader maintains the relative spatial position of each word/line/block etc.

The text block entities identified includes attributes in a similar fashion. The line id, word id, word text, line text of every word and line in the text block, along with coordinates for each word and line as well as the text of the individual lines identified in the block is populated in the database. Additional relationships such as lines below/above a word in the text block are also maintained. For example 'DRAWEE' can be extracted from the document using the line below word relationship. The line below the word DRAWEE in the textblock is 'ABCD PRIVATE LIMITED'.

\section{Incorporation of Universal and Domain specific background knowledge}
It has been observed that most of the documents such as bank receipts, invoices, etc in the real world have certain universal patterns that humans implicitly utilize while reading them. In DeepReader universal background knowledge is incorporated in four different ways:
\begin{enumerate}
 \item Via the use of universal abstract data types such as city, country, date, phone number. This greatly aids extraction as the relevant field can often be retrieved just by looking up the type information.
 \item Via universal validation rules enforcing well known patterns. For example an address must contain a country, city and zip code and every amount or price field must be a real number with an associated currency. These rules help the system validate and refine the resulting extractions.
 \item Via well known aliases or semantically similar entities such as Amount, Amnt, Amt.
 \item Via commonly occuring  visual/text patterns: For example phrases seperated by a colon or different fonts in a box normally constitute a key value pair. In Fig. 4 "SWIFT: SCBLUS33" is an example of a key value pair. Using the key SWIFT its value can be  directly fetched by querying on this relationship table.
\end{enumerate}

In addition to universal background knowledge, domain specific background knowledge is incorporated into the system to correct for OCR spelling mistakes, and for further data typing. For example it may make sense to add a data type called designation for resumes, whereas it may make sense to add a SWIFT number as a data type for financial documents. Users of DeepReader have the flexibility to add this knowledge into their extraction application.

\section{SQL queries to extract useful fields} Once the relational schema is populated with data from the document, it can now be queried like a regular database schema using SQL. Thus the original problem of information extraction from images has been abstracted to the problem of simply querying on a set of tables. This allows for a wider audience that may not be familiar with Vision but is familiar with SQL to specify extraction rules/steps.

%\subsection{Detecting Visual Entities via Vision  APIs}: The document image is initially processed via the DeepReader Vision APIs that help identify visual entities present in the image such as text words, lines, blocks, boxes, tables, and objects. These APIs are hosted on our cognitive API hosting platform Samuhaa \cite{}. The advantage of hosting our vision APIs on Samuhaa is that Samuhaa allows for easy creation of custom APIs, apart from updation via retraining of these vision APIs for a particular use case. This way the platform becomes progressively richer with increased usage resulting in a diverse set of state of the art vision APIs. 

%\subsection{Extracting Relationships between the Visual Entities}: The visual entities extracted above are related to each other in interesting ways and knowledge of these relationships is often useful for information extraction. For example, it may be useful to know that the text in the box with the string "address:"  is "123 Maryland Parkway, MA, USA", or that the text to the right of the string "SWIFT:" is "ABC30221" as shown in Fig. \ref{}. Fig \ref{} outlines the relational schema for capturing relationships in the current version of DeepReader. The relationships thus captured are stored in an unnormalized relational database. The idea is to allow for redundancy in the schema to enable simpler queries for extracting data from the database. 

\section{\textbf{\large Natural Language Interface to Deep Reader }}

While querying in SQL certainly makes the information extraction solutions
more accessible to a wider range of programmers, it may be possible to reach an even wider audience if the system had a conversational interface, obviating the need for familiarity with SQL. There has recently been good progress in converting natural language queries into SQL ~\cite{yu2018typesql} using sequence to sequence models. Deep Reader builds on this work to offer users a conversational interface that maps NL utterances internally into an SQL query, and extracts the relevant information.

While sometimes a simple SQL query will be sufficient to fetch the required information from the database. Many times, a complex query or a sequence of simple queries has to be executed in a pipeline to get the correct information. These complex queries can be broken up into multiple simple queries, storing and building upon the intermediate results. For this to happen, the intermediate results are saved and fed as a data table into the subsequent query. This way a sequence of queries will get executed and result in a workflow which can be saved and applied on other similar documents in one shot to get the required information. This is explained in more detail in the subsequent sections.

\subsection{\textbf{\large Intent Identification} }
The user when interacting with Deep Reader through Natural Language, can ask for a variety of different fields. An intent identifier is necessary to classify the NL-Query and ascertain the intention. We will categorize the given NL-utterances into 3 classes. These are 1) extraction queries, 2) request for creating or saving a workflow \& 3) book-keeping. Once the intent is known, the NL-Query is passed to its respective model for further processing. All the three models are explained in further sections.

We formulate the intent identification as a classif\title{•}ication problem with each NL-utterance being fed as input to a sequential neural model with a softmax output over the possible intents. Recurrent Neural Networks (RNN) \cite{mikolov2010recurrent} offer state of the art results for finding patterns in sequential data. We use Long Short Term Memory (LSTMs) \cite{hochreiter1997long} which is a flavour of RNN and captures important experiences that have very long time lags in between.

\subsection{\textbf{\large Processing Extraction Queries} }
Once the intent identifier classifies the given NL-Query as an extraction query, an NL-Query is sent to this model for further processing. The corresponding SQL-Query is structured as follows:

\begin{center}
\textbf{SELECT} \textit{\$SELECT\_COL} \textbf{FROM} \textit{\$TABLE} \\
\textbf{WHERE} \textit{\$COND\_COL} \textit{\$OP} \textit{\$COND\_VAL}
\end{center}

To map an NL utterance to an SQL query we need to perform the following steps:
\begin{enumerate}
\item \textbf{Entity Recognition:} \$COND\_VAL as shown above is an entity which is document specific and the same has to be extracted from the NL sentence. This corresponds to the Named Entity Recogntion (NER) problem and here we  utilize Conditional Random Fields (CRFs)\cite{lafferty2001conditional}  to solve this problem.

During training, each word in the NL-query is tagged as either an entity or a non-entity and a CRF is trained on this encoding. Once we get the \$COND\_VAL, using the CRF, the same will be replaced with a standard word in the dictionary. For example, 'SWIFT' will be identified as an entity by CRFs in the sentence "Please get me the word towards right of SWIFT" and will be replaced with "Please get me the word towards right of \textless COND\_VAL\textgreater". This will help in processing the NL-query by subsequent models in the pipeline \\

\item \textbf{Template Mapping} We employ a template-based approach to the generation of SQL-Queries and formulate it as a slot filling problem. All simple NL-queries will be mapped to one of the templates in our template pool. We formulate this as a classification problem with the modified NL-Query being classified by a deep sequential model. Below are a few sample SQL templates used.\\

\begin{center}
\small SELECT * FROM {\color{blue}{TABLE}}	\\
WHERE id=(SELECT id FROM {\color{blue}{TABLE}} WHERE string="{\color{red}{VALUE}}")	\\
\end{center}

\begin{center}
\small SELECT * FROM {\color{blue}{TABLE}}	WHERE primary\_str="{\color{red}{VALUE}}"	\\
\end{center}

\begin{center}
\small SELECT SUBSTR( line, pos({\color{red}{VALUE}}),  ) FROM {\color{blue}{TEMP}}	\\
\end{center}

\begin{center}
\small SELECT SUBSTR( line, pos({\color{red}{VALUE1}}), pos({\color{red}{VALUE2}})-pos({\color{red}{VALUE1}}) ) FROM {\color{blue}{TEMP}}
\end{center}

\item \textbf{Table Mapping}
Once the correct template is identified, slots for TABLE and VALUE are required to be filled. The VALUE is readily obtained from the Entity Recognition model. The NL-Query has words with many linguistic variants which can map to the same relevant table. For example, the sentences "get me the word towards the right of SWIFT" and "get me the word immediately next to SWIFT"  will map to the same table "rightof". This mapping is done using an LSTM model trained to classify on these variations.

\end{enumerate}

\begin{table}[h!]
\centering
\begin{tabular}{ | c | c | c | c |}
 \hline
\thead{Model} & \thead{No. of training \\ samples} & \thead{No. of testing \\ samples} & \thead{Accuracy}\\
 \hline
 \makecell{Intent \\ Identification}   &   250  & 31 &   100\%\\
 \hline
 \makecell{Entity \\ Recognition}   &   130  & 130 &   96.3\%\\
 \hline
  \makecell{Template \\ Mapping}  &   160  & 44   & 90.2\%\\
 \hline
 \makecell{Table \\ Mapping}  &   160  & 44   & 100\%\\
 \hline
\end{tabular}
 \caption{Results for different LSTM Models}
 \label{LSTM_results}
\end{table}

\subsection{\textbf{\large Creating a Workflow: Sequence of extraction queries} }
Simple extraction queries will only fetch information which is readily available from the database. Often complex queries need to be executed to extract relevant information. Designing complex queries for every possible use case would blow up the SQL-template space and would inhibit query reuse. However, complex queries can be broken down into multiple simple queries, allowing for storing and building upon the intermediate results.

%\begin{center}
%SELECT * FROM {\color{blue}{TEMP}} WHERE primary\_str="{\color{red}{VALUE}}"	\\
%\end{center}

For example, as shown in the figure \ref{fig:sample_city} , to get the "Account" information, below are the set of NL-Queries needs that to be executed in a sequence.

\begin{itemize}
\item "Kindly get the block information for the block containing the word remit"
\item "Please get the line which has word Account in it from the previous result"
\item "Get substring which is towards right of Account from the previous result"
\end{itemize}

Different combinations of simple queries executed in sequence will fetch the complex entity. By default, the output of any intermediate result is stored in a temporary table which can be queried further.

\begin{figure*}[ht]
\centering 
\includegraphics[scale=0.3]{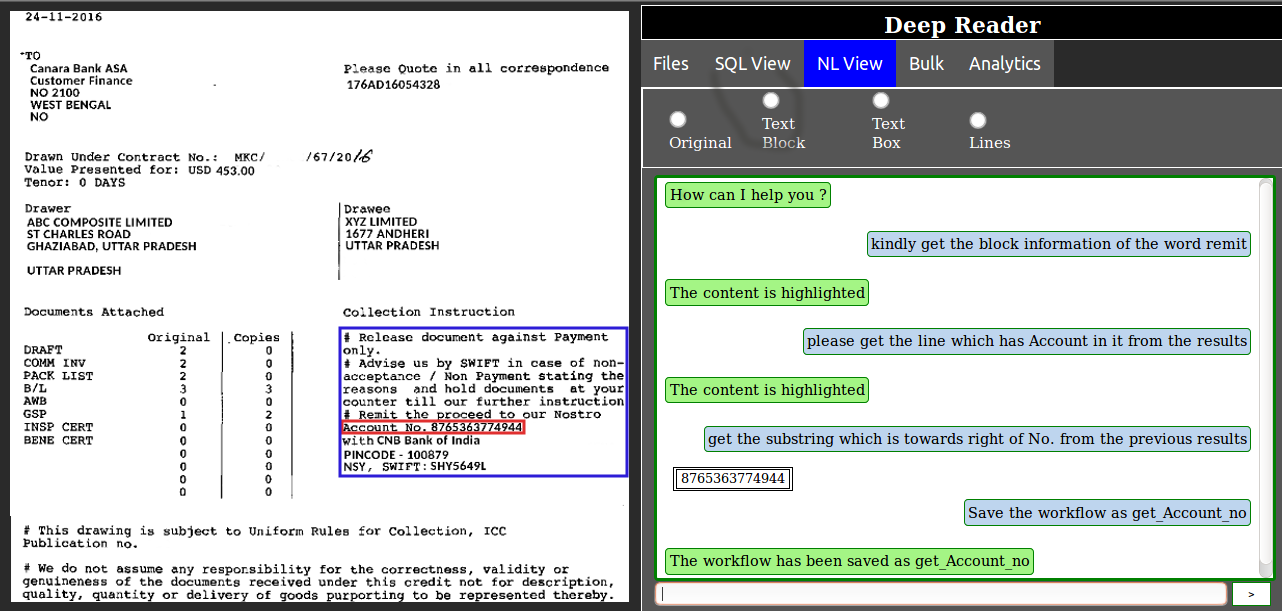}
\caption{Graphical User Interface}
\label{schema}
\end{figure*}

\subsection{\textbf{\large Book-Keeping} }
A sequence of meaningful NL-Queries will fetch the required information. Once saved, this workflow can be applied to a new document with a similar template. A simple framework using queues has been built in order to store the recent set of commands in a sequence. Once the user fetches a meaningful information, the workflow is saved. Simple NL-Queries like "clear the workflow", "save the workflow", "apply the workflow on this document" etc.. are used for book-keeping.

%===========================================================

%------------------------------------------------------------------------- 
\section{Conclusions and Future Work}
This paper introduces DeepReader, a framework for the extraction of relevant data from image documents.  We have elaborated on the vision challenges involved in processing real documents such as removal of background noise and hand written text recognition.  We proposed a GAN based architecture for noise removal and a state of the art sequence-to-sequence attention model for handwritten text recognition. In addition, the spatial relationships between visual entities are captured via a rich predefined relational schema.  A Siamese Network was used to identify the template of a document. A NL Interface has been integrated to allow novice users to converse with the system for rapid and easy data extraction. 

The current DeepReader framework can be used to process a wide variety of real world documents that adhere to a finite number of templates, however several enhancements are required for better generalization. Organizing universal background knowledge by developing a framework for prioritizing rules for a particular context, and allowing for easy inter-operability to domain ontologies for easy integration of domain specific knowledge are part of the future plans. An aspect that is being explored is learning the rules for extraction automatically via examples. This would allow users to specify the extraction fields automatically without writing down elaborate extraction rules. The underlying philosophy is that while the innate human ability to see and discriminate objects in the real world is not yet fully understood, the ability to read  and understand documents can be performed via a combination of background knowledge and visual/textual cues. With the progress of deep learning in being able to capture the visual elements more accurately, perhaps document understanding can now be accomplished to a much more satisfactory degree. 
%============================================================
\pagebreak
\bibliographystyle{splncs}
\bibliography{egbib}
\end{document}